\documentclass[11pt]{article}

\usepackage[preprint]{acl}

\usepackage{times}
\usepackage{latexsym}
\usepackage[T1]{fontenc}
\usepackage[utf8]{inputenc}
\usepackage{microtype}
\usepackage{inconsolata}
\usepackage{graphicx}
\usepackage{booktabs}   
\usepackage{amsmath}
\usepackage{algorithm}
\usepackage{algpseudocode}
\usepackage{amsfonts}
\usepackage{enumitem} 

\setlist{nosep} 
\title{Decoding-Level Taboo: A Diagnostic Stress Test for LLM Robustness}

\author{%
  Tadanobu Chuyo Kamijo \\
  University of the Ryukyus \\
  \texttt{tadanobu@cs.u-ryukyu.ac.jp}
  \And
  Ori Rottenstreich \\
  Technion \\
  \texttt{or@technion.ac.il}
  \AND
  Javier Conde \qquad Gonzalo Martínez \qquad Pedro Reviriego \\
  Infornation Processing and Telecommunications Center (IPTC) \\
  Universidad Pol{\'e}cnica de Madrid \\
  \texttt{\{javier.conde.diaz, gonzalo.martinez.ruizdearcaute, pedro.reviriego\}@upm.es}
}

\begin{document}
\maketitle

\begin{abstract}

Large language model evaluations typically focus on performance under nominal conditions, creating an illusion of capability where models comfortably walk a narrow, highly optimized generation corridor. In real-world deployments, however, complex system prompts, safety guardrails, and structural constraints continuously force models off this nominal path, driving a divergence between benchmark scores and deployment performance. To address this issue, we introduce Decoding-Level Taboo, a zero-prompt diagnostic stress test that intervenes directly in logit space at runtime, forcing models out of their nominal paths. By dynamically masking primary candidate tokens at word boundaries, Taboo forces machine circumlocution. 

Evaluating Taboo across several open-weight model families reveals that off-path robustness is heavily influenced by both parameter scale and post-training instruction alignment, with robustness generally improving with model size and alignment. Beyond the results presented in this paper, Taboo provides a novel primitive for generating diverse synthetic datasets, stress-testing runtime safety guardrails, and auditing model reliability prior to real-world deployment.

\end{abstract}

\section{Introduction}

Evaluating Large Language Models (LLMs) focuses primarily on decoding under nominal conditions, assessing the capability along narrow paths of preferred token sequences \cite{wei2021finetuned, polo2024tinybenchmarks}. While models achieve impressive scores across standard leaderboards, this nominal paradigm creates an illusion of capability: it obscures whether a model possesses adaptable, robust internal reasoning or merely relies on superficial memorized trajectories.

In production deployments, however, models rarely operate in unconstrained environments. Complex system instructions, safety guardrails, structured JSON output formats, and negative constraints continuously put pressure on generation \cite{ouyang2022training, rafailov2024direct}. These real-world operational demands force models off their nominal decoding corridors. Despite the ubiquity of these constraints, it remains fundamentally unclear how an LLM's internal multi-step reasoning engine holds up under continuous off-path stress. Specifically, whether the model dynamically re-routes its latent reasoning or suffers logic degradation. For instance, when presented with a standard grade-school math problem from GSM8K, Llama-3.1-8B-Instruct effortlessly reaches the correct answer under greedy decoding; yet, blocking its most-preferred token at each word start, the decoding-level intervention we formalize in Section~\ref{sec:taboo}, frequently causes the model's reasoning chain to unravel, triggering repetitive token loops or erroneous arithmetic\footnote{See Appendix~\ref{app:failures}.}. Despite the underlying mathematical logic remaining identical, subtle constraints force the model off its typical generation path, exposing its fragility.

Recent efforts to evaluate non-nominal generation, such as IFEval \cite{zhou2023instruction} and FollowBench \cite{jiang2024followbench}, attempt to address this by introducing negative constraints (e.g., forbidding specific keywords). While these benchmarks highlight compliance challenges, they suffer from two core limitations. First, they evaluate simple, isolated token rules rather than continuous structural stress during multi-step generation \cite{vocabbans2026}. Second, because they operate strictly at the prompt level, they inherently confound rule-following compliance with internal reasoning resilience \cite{candussio2026dontsayit, ayoobi2026tokenizer}. A failure on a prompt-based ban cannot distinguish between a model refusing to obey an instruction and a complete collapse of its underlying logic engine when forced off-path.

This fundamental limitation underscores the need for diagnostic frameworks that bypass the prompt interface entirely. Modifying input prompts introduces significant confounding variables, including instruction-parsing overhead, surface-level sensitivity to minor phrasing variations \cite{zhao2021calibrate, sclar2023quantifying}, and attention allocation drift across extended context windows. To rigorously measure whether an LLM's latent reasoning engine remains resilient when forced off its primary decoding trajectory, diagnostic evaluations must decouple the stress perturbation from the prompt. Operating directly within the generation pipeline via runtime logit-space intervention enables precise, unconfounded probing of model health.

In this paper, we introduce Decoding-Level Taboo: a zero-prompt diagnostic stress test that intervenes directly within the autoregressive decoding loop at runtime. Holding the prompt fixed, Taboo dynamically masks primary candidate tokens at word start boundaries, preventing the model from traversing its most confident, highly optimized generation path. This intervention forces the model into machine circumlocution, requiring its latent reasoning engine to dynamically construct semantically valid, off-path alternatives on the fly. We parameterize the severity of this off-path stress using the injected surprisal, which measures the distributional shift induced when a model is diverted from its preferred decoding corridor. By isolating off-path execution from prompt interpretation, Taboo provides a clean diagnostic lens into whether an LLM's multi-step reasoning remains structurally intact under forced divergence.

The main contributions of this work are:

\begin{enumerate}
    \item We propose \emph{Decoding-Level Taboo}, a zero-prompt, runtime diagnostic stress test that intervenes directly in logit space at word boundaries to quantify a model's off-path reasoning resilience.

     \item Using the proposed Taboo decoding, we show across four open-weight families and four benchmarks that off-path robustness is governed jointly by scale and post-training alignment: aligned checkpoints absorb a \emph{larger} effective perturbation yet retain far more of their multi-step reasoning. An effect that widens with scale, is family-dependent (absent in Llama-3 at 7--8B, though it reappears by 70B), task-specific (generative reasoning, not MCQ or factual recall), and bounded by formal syntax (HumanEval collapses to zero).

    \item We position Taboo decoding as a versatile, zero-training runtime primitive beyond diagnostic evaluation, discussing its potential across safety auditing, synthetic Chain-of-Thought (CoT) trajectory discovery, structured output stress-testing, and taboo-guided policy alignment.
     
\end{enumerate}

\section{Decoding-Level Taboo}
\label{sec:taboo}

In this section, we formalize the proposed Decoding-Level Taboo, a zero-prompt diagnostic framework designed to evaluate LLM reasoning resilience under off-path decoding stress.

\subsection{General Framework and Logit Interventions}

Let an autoregressive language model parameterize a probability distribution over a vocabulary $V$ given a prefix sequence $x_{<t} = (x_1, x_2, \dots, x_{t-1})$. At each decoding step $t$, the model's neural architecture produces a vector of unnormalized logits $z_t \in \mathbb{R}^{|V|}$. Under nominal greedy decoding, the model deterministically selects the top candidate token $x_t^{\text{nom}} = \arg\max_{j \in V} z_{t,j}$, following its most confident, memorized generation path.

To force off-path generation without altering the input prompt, Decoding-Level Taboo applies an additive binary logit mask $M_t \in \{0, -\infty\}^{|V|}$ to the logits before applying the softmax operator:

\begin{equation}
P_{\text{taboo}}(x_t \mid x_{<t}) = \text{softmax}\left(z_t + M_t\right)
\end{equation}

By setting $M_{t,j} = -\infty$ for the top-scoring candidate indices $\mathcal{K}_t = \text{Top-}i(z_t) \subset V$ (the $i$ most probable nominal tokens), we set their conditional generation probabilities to zero. This intervention removes the model's primary decoding preferences, forcing its latent reasoning engine into machine circumlocution and constructing semantically coherent, syntactically valid alternatives.

\subsection{Word-Initial Intervention Strategy}

Applying logit masking indiscriminately at every subword step severely corrupts tokenization structure, forcing models into invalid byte sequences or repetitive character fragments. To isolate semantic reasoning from tokenization corruption, Taboo restricts interventions exclusively to word-initial token steps, leaving mid-word subword continuations entirely unmasked.

When generating an autoregressive sequence, each decoding step $t$ either initiates a new lexical word or continues an existing subword fragment. We define a binary indicator predicate $\mathcal{W}(x_t, x_{<t}) \in \{0, 1\}$ that evaluates to $1$ when the token carries a tokenizer word-boundary marker, an approach that can miss the first generated token, punctuation-attached words, and numerals, and $0$ otherwise. The additive logit mask $M_{t,j}$ applied to candidate token $j$ at step $t$ is defined as:

\begin{equation}
\small
M_{t,j} = 
\begin{cases} 
-\infty & \text{if } \mathcal{W}(x_t, x_{<t}) = 1 \text{ and } j \in \text{Top-}i(z_t) \\
0 & \text{otherwise}
\end{cases}
\end{equation}

\noindent where $\text{Top-}i(z_t)$ denotes the indices of the $i$ highest unnormalized logits at step $t$; we refer to $i$ as the \emph{taboo rank} (mask width). Word-initial status is a static, vocabulary-level property precomputed once as a boolean mask based on standard tokenizer word-boundary markers (\.{G} in byte-level BPE, \textvisiblespace\ in SentencePiece). $\mathcal{W}$ is evaluated on the \emph{nominal top candidate} at step $t$: an intervention is triggered only when the model's preferred continuation would start a new word. This setup establishes a two-phase decoding lifecycle for every word:
\begin{enumerate}
    \item \textbf{Word Initiation ($\mathcal{W}=1$):} When the nominal top candidate is word-initial, the top $i$ candidates are masked ($M_{t,j} = -\infty$), forcing greedy decoding to select the next-ranked token $x_t^\ast = \arg\max_{j \notin \text{Top-}i(z_t)} z_{t,j}$.
    \item \textbf{Subword Completion ($\mathcal{W}=0$):} For all subsequent token steps needed to complete that word, masking is completely disabled ($M_t = 0$). The model decodes its chosen subword continuation fragments without interference.
\end{enumerate}

In a controlled ablation on GSM8K (with $n=100$ questions), applying Taboo masking at \emph{mid-word} positions (non-start subwords) while leaving word starts unmasked collapses absolute accuracy to $0.01$--$0.05$ (at matched mask width $i=2$) across all four model families and both BPE and SentencePiece tokenizers. For instance, under mid-word masking at a modest average intervention Gemma-3-12B-it falls to $0.01$ accuracy; in contrast, word-initial masking at a much higher intervention preserves $0.81$ absolute accuracy. Intervening at word starts applies semantic stress, whereas intervening mid-word merely induces tokenization corruption.

As a further check that the measured drops reflect disrupted reasoning rather than the intervention mechanically colliding with a task's own constraints, we evaluated the \texttt{forbidden\_words} instruction type from IFEval, where a designated word must be avoided. Individual generations do change under Taboo, the intervention alters the decoding path, so per-item compliance flips in both directions, but taboo cannot mechanically force a forbidden word to appear, and \emph{aggregate} compliance does not degrade in any family (net change $+0.0$ to $+0.2$; $n=25$ prompts, indicative). The drops elsewhere are therefore not explained by the intervention systematically breaking the task constraint.

Even with interventions restricted strictly to word starts, Taboo reveals a fundamental divide between natural language and formal domain synthesis. While natural language possesses rich semantic redundancy, allowing models to leverage synonyms and circumlocution when primary word-initial tokens are blocked, formal code exhibits near-zero lexical redundancy at structural word boundaries. When Taboo masks a mandatory word-initial keyword or operator (such as \texttt{def}, \texttt{return}, or \texttt{for}), the model is deprived of functional alternatives. Because formal syntaxes offer no valid off-path word choices to express the same programmatic intent, forcing divergence at word starts induces immediate structural breakdown. Empirically, word-initial Taboo decoding on HumanEval \cite{chen2021evaluating} drives all evaluated models to near-zero Pass@1 accuracy (every cell $0\%$ except a single $1/102$ retention for Llama-3.1-8B-Instruct at $i{=}2$), indicating that formal syntactic constraints establish an effective lower bound for off-path generation.

\subsection{Quantifying Stress via Injected Surprisal}

To systematically measure the severity of the intervention required to divert the model off-path, we define Injected Surprisal ($\Delta S_t$). Injected surprisal measures the instantaneous information-theoretic cost (in bits) introduced by masking the nominal candidate set at step $t$:

\begin{equation}
\begin{split}
\Delta S_t &= -\log_2 P_{\text{nom}}(x_t^\ast \mid x_{<t}) - \\
           &\quad \left( -\log_2 P_{\text{nom}}(x_t^{\text{nom}} \mid x_{<t}) \right)
\end{split}
\end{equation}

\noindent where $x_t^\ast = \arg\max_{j \in V} (z_{t,j} + M_{t,j})$ is the selected off-path token, and $P_{\text{nom}}$ represents the unconstrained nominal distribution. Both terms are evaluated under the \emph{nominal} (unmasked) distribution, so $\Delta S_t \geq 0$ by construction: it is the extra surprisal, in the model's own belief, of emitting the forced alternative instead of its preferred token.

If step $t$ is not a word boundary, or if no intervention is applied, $M_t = 0$, yielding $\Delta S_t = 0$. For a generated sequence of length $T$, the total accumulated off-path stress dose $\mathcal{S}_{\text{total}}$ is defined as:

\begin{equation}
\mathcal{S}_{\text{total}} = \sum_{t=1}^{T} \Delta S_t
\end{equation}

Normalizing $\mathcal{S}_{\text{total}}$ by the total number of word-boundary interventions provides a standardized metric, the \textit{Mean Injected Surprisal Per Intervention} ($\bar{\Delta S}$), allowing direct comparability of decoding stress across different model architectures, parameter sizes, and tokenizers.

\subsection{Implementation}

We implement Decoding-Level Taboo as a modular pipeline component within the Hugging Face transformers library. Because modern LLM tokenizers employ distinct byte-level or subword encoding schemes (e.g., Byte-Pair Encoding in Llama-3 and Qwen2.5 vs. SentencePiece in Gemma-3), word boundary identification relies on tokenizer-specific prefix metadata. Specifically, we precompute a vocabulary-sized boolean mask marking every token whose decoded string begins with the tokenizer's word-boundary prefix, and consult it for the nominal top candidate at each step.

At each decoding step $t$, the TabooLogitsProcessor accepts the generated token matrix \texttt{input\_ids} and the unnormalized logit tensor \texttt{scores} $\in \mathbb{R}^{B \times |V|}$ for batch size $B$. For each sequence in the batch, the processor evaluates whether step $t$ corresponds to a word boundary using tokenizer prefix metadata. If a word boundary is identified, it retrieves the top-$i$ candidate token indices, masks their corresponding logits to $-\infty$, and selects the highest-scoring unmasked token as the off-path output $x_t^\ast$. It then computes the step surprisal $\Delta S_t$ as the difference in log-probability between the unconstrained nominal top candidate and the newly selected off-path token. If step $t$ represents an intermediate subword continuation ($\mathcal{W}=0$), the logit tensor remains unmodified.

Because the intervention is a vectorized top-$i$ selection and masked argmax applied only at word-initial steps, its per-step cost is negligible relative to the transformer forward pass; the only material cost is indirect, blocking the preferred continuation lengthens generations (mean generated tokens grow by ${\approx}1.4\times$ at $i{=}1$ on GSM8K, see Appendix~\ref{app:token_expansion}). Tracking and logging $\Delta S_t$ in parallel with generation enables dynamic monitoring of off-path stress trajectories without requiring multi-pass generation or backwards gradients.

\section{Experimental Setup}

To evaluate how off-path decoding stress affects model performance across diverse reasoning modalities, we construct a controlled evaluation environment comparing base and instruction-tuned models across four distinct benchmark tasks\footnote{The code is available at https://doi.org/10.5281/zenodo.21761445 (anonymous)}.

\subsection{Model Suite}
\label{sec:model_suite}

We evaluate major open-weight model families spanning parameter scales from 0.5B up to 72B parameters:
\begin{itemize}
    \item \textbf{Qwen2.5 Family} \cite{qwen2_5}: Multilingual models with robust nominal reasoning across code and mathematics.
    \item \textbf{Gemma-3 Family} \cite{gemma3_2025}: Google's open-weights models with distinct pre-training and alignment optimizations.
    \item \textbf{Llama-3 Family} \cite{dubey2024llama}: Meta's widely adopted dense transformer benchmark models.
    \item \textbf{OLMo-2 Family} \cite{olmo2024}: Fully open-source pre-trained and aligned models with transparent pre-training data mixtures and training recipes.
\end{itemize}

\noindent To isolate the structural impact of post-training alignment (e.g., SFT, DPO, RLHF) versus raw pre-training capability, we evaluate both the Base (pretrained) and Instruct/Aligned checkpoints for each architecture.

All models were run in 4-bit (bitsandbytes nf4, bf16 compute); the small-to-mid ladder was additionally run in native \texttt{bf16} as a precision control. Across Qwen2.5-\{3B,7B,14B\}, OLMo-2-7B and Llama-3.1-8B (base and instruct, GSM8K, on the same seeded items), the median absolute difference $|\mathrm{acc}_{\text{4-bit}}-\mathrm{acc}_{\text{bf16}}|$ over the 50 (model, dose) cells is $0.03$ and 47 of 50 are $\le 0.07$; the single larger discrepancy is Qwen2.5-14B-Instruct at $i{=}2$ ($0.62$ bf16 vs.\ $0.44$ 4-bit). Re-running that model in 8-bit places this cell at $0.58$, matching bf16 rather than 4-bit, so the anomaly is specific to aggressive 4-bit quantization. All qualitative conclusions (the instruct-over-base ordering, the dose-response shape, and the Llama exception) reproduce across precisions, so quantization does not drive the findings.

\subsection{Evaluation Benchmarks}

Our suite spans distinct cognitive and structural domains to contrast semantic flexibility against rigid syntactic constraints:
\begin{itemize}
    \item \textbf{GSM8K} \cite{cobbe2021gsm8k}: multi-step grade school mathematical word problems requiring chained algorithmic logic.
    \item \textbf{MMLU (Analysis of Format Constraints)} \cite{hendrycks2021mmlu}: Four-way multiple-choice questions spanning 57 knowledge domains, probing factual and conceptual recall under a fixed answer format. Because the single-letter answer token is inherently word-initial, the intervention directly masks the target choice rather than forcing semantic circumlocution. We intentionally include this benchmark to illustrate the structural boundaries of decoding-level constraints.       
    \item \textbf{TriviaQA} \cite{joshi2017triviaqa}: open-domain factual recall and knowledge retrieval without chain-of-thought dependencies.
    \item \textbf{HumanEval (Sanity Check / Negative Control)} \cite{chen2021evaluating}: Python function synthesis. As established in Section 2, because formal programming languages lack semantic redundancy at keyword boundaries, HumanEval serves as a negative control.
\end{itemize}

\subsection{Inference Procedure and Extraction}

To establish a uniform baseline across base and instruct variants, all evaluations adhere to a standardized 2-shot prompting setup. This context provides basic structural anchors without introducing task-specific shortcut solutions. We employ CoT prompting throughout, extracting the final answer from the model's completed generation. Each experimental configuration is evaluated on a fixed, seeded subsample of the benchmark's test split ($n=500$ for the primary $7\text{--}12\text{B}$ comparisons; $n=100$ for the scale ladder and $\ge 32\text{B}$ runs), ensuring identical test items are used across all conditions for a given model.


All conditions are scored deterministically. For GSM8K we extract the final number (an explicit \texttt{Answer:} line if present, else the last number) and compare numerically; for TriviaQA we apply SQuAD-style normalization and match the reference alias set; for MMLU we extract the answer letter and for HumanEval we execute the completed function against the reference tests in a sandboxed subprocess (Pass@1). To confirm that deterministic extraction is not itself the source of the observed drops, we ran a sanity check against an LLM judge (\texttt{Gemini 3.6 Flash}) on a 200-item subsample: on GSM8K the final-number extraction agreed with the judge on $97\%$ of items ($100\%$ on taboo-perturbed outputs); the residual MMLU disagreements (agreement $79\%$) trace to few-shot models over-generating a hallucinated follow-up question, which the deterministic first-answer extraction handles but which misleads the judge. Because strict extraction can only undercount a mangled-but-correct answer, our retention estimates are conservative.

\subsection{Evaluation Metrics}

To systematically evaluate model performance under nominal and forced off-path conditions, we report two baseline metrics and introduce a conditional retention metric:

\begin{enumerate}
    \item Nominal Accuracy ($Acc_{\text{Base}}$): The standard benchmark accuracy under unconstrained, greedy decoding ($M_t = 0$ at every step), evaluated across the full dataset $\mathcal{Q}$.
    \item Taboo Accuracy ($Acc_{\text{Taboo}}$): The overall model accuracy when subjected to Decoding-Level Taboo intervention at mask width $i$.
\end{enumerate}

To isolate intrinsic off-path reasoning resilience from nominal model capability, and to eliminate confounding noise from stochastic recovery on base-incorrect items, our primary metric is the Conditional Retention Ratio ($\mathcal{R}$). We restrict $\mathcal{R}$ to the subset of instances $\mathcal{Q}_{\text{correct}} \subseteq \mathcal{Q}$ where the model successfully solves the problem under nominal greedy decoding:

\begin{equation}
\small
\mathcal{R} = \frac{1}{|\mathcal{Q}_{\text{correct}}|} \sum_{i \in \mathcal{Q}_{\text{correct}}} \mathbb{I}\left(\text{Eval}(\hat{y}_i^{\text{Taboo}}) = y_i\right) \times 100\%
\end{equation}

\noindent where $y_i$ is the ground-truth target and $\hat{y}_i^{\text{Taboo}}$ is the generated output under Taboo constraints. A conditional retention of $\mathcal{R} = 100\%$ indicates total reasoning immunity under off-path stress, whereas $\mathcal{R} = 0\%$ signifies complete logic collapse when forced off the model's nominal paths. 

\section{Empirical Results and Analysis}

\begin{figure*}[t]
  \centering
  \includegraphics[width=\textwidth]{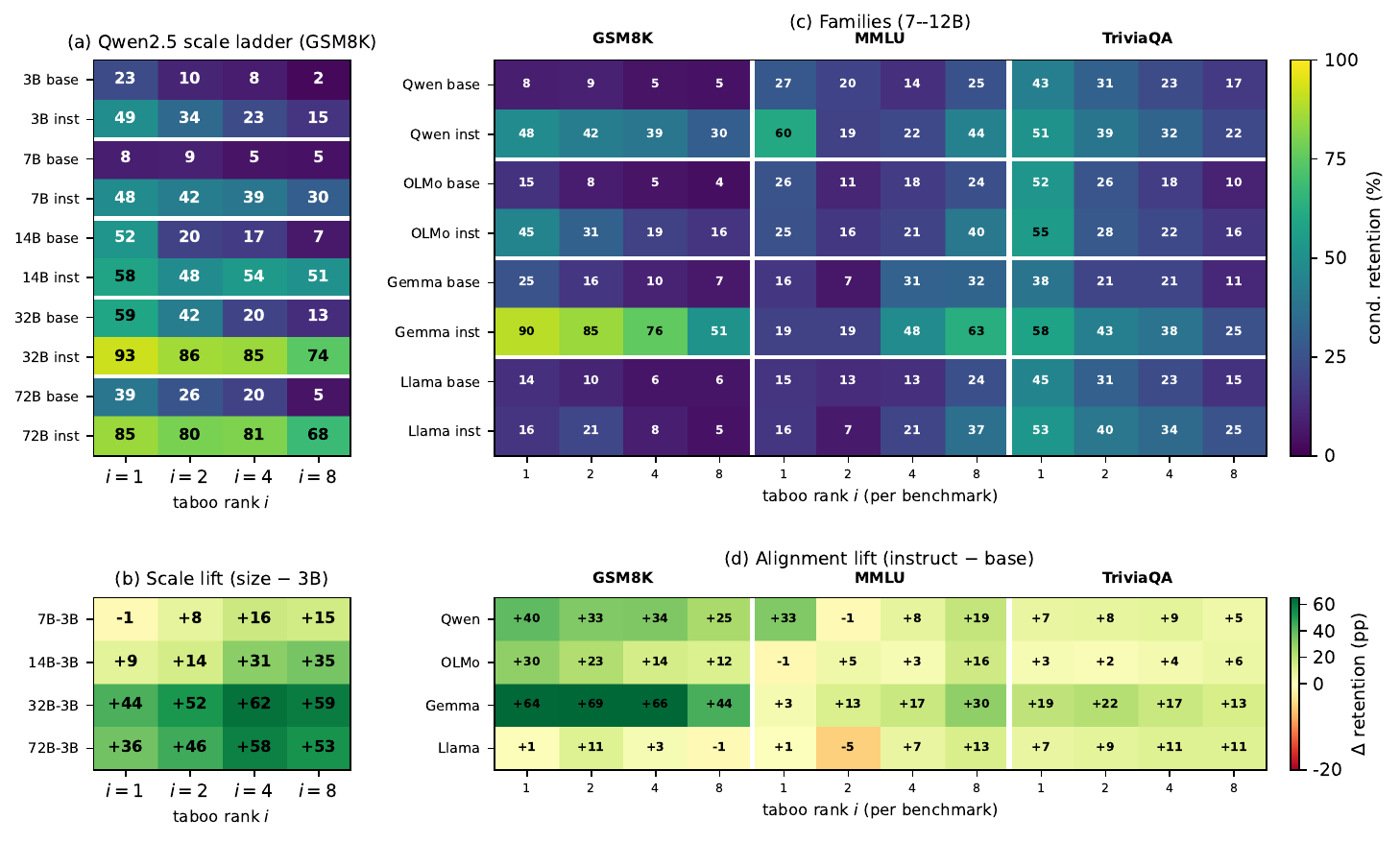}
  \vspace{-8mm}
  \caption{\textbf{Alignment and scale both confer off-path robustness; the alignment effect is task-specific.} Conditional retention $\mathcal{R}$ (percent of baseline-correct items still correct under Taboo) as annotated heatmaps; within each block, columns are the taboo rank $i\in\{1,2,4,8\}$. 
  \textbf{Left (GSM8K, the clean scale axis).} \textbf{(a)} the Qwen2.5 scale ladder (3B--72B), base and instruct rows adjacent, so alignment reads vertically and scale reads down the panel; \textbf{(b)} scale lift, instruct retention at each Qwen size minus the smallest (3B). 
  \textbf{Right (four families at 7--12B, across three benchmarks).} \textbf{(c)} absolute retention for GSM8K, MMLU and TriviaQA; \textbf{(d)} alignment lift (instruct minus base) per benchmark. The effect is large and clean on GSM8K (Gemma/Qwen/OLMo, absent in Llama-3 at this size), noisy and non-monotone on MMLU (an MCQ-floor artifact), and small on TriviaQA, specific to generative multi-step reasoning. Panels (a,c) share the sequential scale; (b,d) the diverging scale (pp).}
  \label{fig:heatmap}
  \vspace{-4mm}  
\end{figure*}

We evaluate the reasoning resilience of open-weight LLMs by measuring their performance under Decoding-Level Taboo across our benchmark suite. Figure~\ref{fig:heatmap} summarizes the results showing the Conditional Retention Ratio $\mathcal{R}$ as a function of the taboo rank $i$, contrasting Base and Instruction-tuned checkpoints and model size for the Qwen family. HumanEval is omitted from the grid as all models score near-zero ($\mathcal{R}\approx0\%$, a single $1/102$ cell aside), serving as an off-grid negative control. To provide a complete characterization of these dynamics, we relegate extended granular views to the appendices: (1) retention trends for the 7--12B model size regime are plotted as curves in Appendix~\ref{app:results_CI}, overlaid with 95\% Wilson score intervals computed from the baseline-correct sample size $|\mathcal{Q}_{\text{correct}}|$ to visualize the item-sampling uncertainty behind the observed differences; and (2) absolute accuracy across all evaluated model configurations is detailed in Appendix~\ref{app:results_full_lader}.

\subsection{Base vs. Instruction-Tuned Models}
\label{sec:base_vs_instruct}

The clearest signal in our diagnostic is that instruction tuning confers off-path robustness, but only for tasks that require multi-step reasoning, and only in some families. On GSM8K, aligned checkpoints retain a far larger fraction of their baseline-correct items under Taboo than their base counterparts. At the mildest dose ($i=1$), conditional retention rises from $8\%$ to $48\%$ in Qwen2.5-7B, from $15\%$ to $45\%$ in OLMo-2-7B, and from $25\%$ to $90\%$ in Gemma-3-12B. The gap persists, and in Gemma even widens in relative terms, as the dose grows: at $i=8$ the instruct checkpoints still retain $30\%$ (Qwen), $16\%$ (OLMo) and $51\%$ (Gemma), while every base model has collapsed below $8\%$.

This alignment benefit is neither universal nor uniform. Llama-3.1-8B shows essentially no separation between base and instruct on GSM8K ($14\%$ vs.\ $16\%$ at $i=1$, both falling to $\sim\!5\%$ by $i=8$), a consistent exception across doses. And the benefit is task-specific: on TriviaQA the base--instruct gap is small in every family (e.g.\ $53\%$ vs.\ $45\%$ for Llama at $i=1$), while on MMLU both checkpoints are weak and non-monotone, an artifact of the multiple-choice floor, where forced re-selection can raise retention at large $i$. The effect is thus specific to generative multi-step reasoning: alignment appears to teach models to reconstruct a valid reasoning path once their preferred tokens are removed, rather than to recall facts or refill MCQ slots more robustly.

\subsection{Parameter Scale vs. Off-Path Resilience}

Scale generally improves off-path robustness on generative reasoning, though not monotonically, and the effect compounds with alignment (Figure~\ref{fig:heatmap} a,b). On the Qwen models, for GSM8K, conditional retention at the mildest dose ($i=1$) rises with size for base checkpoints overall, $6\%$ (0.5B), $11\%$ (1.5B), $23\%$ (3B), $52\%$ (14B), $59\%$ (32B), though not monotonically: 7B base dips to $8\%$. Their aligned counterparts are markedly more robust, from $6\%$ at 0.5B to $93\%$ at 32B and $85\%$ at 72B. The alignment gap therefore \emph{widens} with scale: at the top of the ladder an aligned 32B model retains almost everything it could solve ($93\%$ at $i=1$, still $74\%$ at $i=8$), whereas its base sibling drops to $13\%$ at $i=8$. Interestingly, the largest base model is not the most robust: Qwen2.5-72B-base retains less than 32B-base ($39\%$ vs.\ $59\%$ at $i=1$), a non-monotonicity suggesting that raw scale alone does not guarantee off-path stability.

Scale does not rescue every domain. On MMLU, absolute accuracy under Taboo collapses toward the multiple-choice floor at \emph{every} size, including 72B, consistent with the format artifact discussed above rather than a capability that scale could buy back. On TriviaQA, degradation is graceful at all sizes and the base, instruct gap stays small, confirming that the compounding scale$\times$alignment benefit is specific to multi-step generative reasoning.

\subsection{Cross-Family Architectural Divergence}
\label{sec:cross_family}
Holding size roughly constant ($7$--$12$B) and prompting identical, the four families divide cleanly by how much robustness their post-training confers on GSM8K: Gemma-3 gains the most (conditional retention $25\%\!\rightarrow\!90\%$ at $i=1$), Qwen2.5 and OLMo-2 gain strongly ($8\%\!\rightarrow\!48\%$ and $15\%\!\rightarrow\!45\%$), and Llama-3 gains essentially nothing ($14\%\!\rightarrow\!16\%$, confidence intervals overlapping). The injected-surprisal profiles (see Appendix \ref{app:inj_surprisal}) offer a mechanistic hint: in the three families where alignment helps, the aligned checkpoint absorbs \emph{more} surprisal per intervention than its base (consistent with a sharper next-token distribution), whereas Llama-3-Instruct absorbs no more than its base, suggesting its post-training leaves the next-token distribution comparatively flat. Since all families are evaluated under the same intervention, prompting, and scorer, the divergence points to post-training recipes (data mixture, SFT/DPO/RLHF composition) rather than the intervention itself. Identifying which recipe component confers taboo-robustness is an open question our diagnostic makes measurable. This family-level exception is genuine but bounded by scale. Raising Llama-3.1-8B's generation budget to 1024 tokens does not recover it, the model still saturates the budget and its retention is unchanged ($22\%$ vs.\ $24\%$ at $i{=}1$), so it reflects a failure to terminate a coherent chain rather than answer truncation. Yet extending the ladder to Llama-3.1-70B, the alignment gain reappears (marginal GSM8K retention rises from $36\%$ base to $75\%$ instruct at $i{=}1$, and $9\%$ to $22\%$ at $i{=}8$): Llama-3 acquires taboo-robustness later in scale than the other three families rather than lacking it, consistent with the scale$\times$alignment compounding above. These are single-run marginal-retention figures at $n{=}100$; with only the 8B and 70B points, we cannot locate the transition.

\subsection{Domain Constraints: Natural Language Redundancy vs. Code Inflexibility}

Comparing performance across benchmark columns highlights the role of domain-specific lexical redundancy (Figure~\ref{fig:heatmap} c):

\begin{enumerate}
    \item \textbf{TriviaQA (Col. 3 - High Redundancy):} Exhibits the highest global retention. Natural language question answering offers vast surface-form flexibility, allowing models to circumvent masked tokens effortlessly.
    
\item \textbf{MMLU (Col. 2 - Multiple Choice):} Performance collapses toward the guessing floor. Because the single-letter answer is inherently word-initial, Taboo directly masks the model's primary selection and forces an arbitrary re-selection. This induces a non-monotonic dose response, showing that MMLU captures a formatting artifact rather than genuine reasoning resilience.
    \item \textbf{GSM8K (Col. 1 - Structured Reasoning):} Demonstrates moderate retention. While mathematical reasoning allows verbal circumlocution, forced rephrasing increases context length and accumulated error probability, leading to eventual logic drift.
    \item \textbf{HumanEval (Negative Control Baseline):} Confirms the empirical bound established in Section 2 without requiring dedicated plot visualization. Masking reserved keywords at word boundaries instantly breaks Python AST construction.
\end{enumerate}

\section{Broader Applications of Decoding-Level Interventions}
\label{sec:applications}

This work introduces Decoding-Level Taboo primarily as a diagnostic tool for internal reasoning resilience, but runtime logit-space intervention can have broader utility as described in the following.


\subsection{Auditing Alignment and Surface-Level Refusals}
Safety alignment paradigms frequently train models to emit canonical refusal templates when presented with adversarial inputs \cite{ouyang2022training, rafailov2024direct}. However, prompt-level evaluations struggle to distinguish whether a refusal reflects deep safety alignment within latent representations or merely a shallow, top-token preference for memorized refusal triggers \cite{candussio2026dontsayit}. By dynamically masking candidate refusal tokens at decoding time, Taboo enables non-destructive safety auditing: if suppressing primary refusal tokens causes the model to immediately emit policy-violating content, the alignment is superficial; if the model instead re-routes to alternative safe phrasings or valid policy rationales, the safety guardrail is structurally robust.

\subsection{Generating Diverse Trajectories for Synthetic CoT Datasets}
Knowledge distillation and reasoning alignment rely heavily on sampling diverse, step-by-step CoT trajectories from teacher architectures \cite{hinton2015distilling, wei2021finetuned}. Standard stochastic sampling techniques, such as temperature scaling or top-$p$ nucleus sampling, often yield repetitive reasoning paths or inject unguided entropy that degrades multi-step logical coherence. By enforcing decoding-level Taboo constraints at word boundaries, the model is systematically diverted from its primary greedy corridor into alternative, logically sound reasoning trajectories. This provides a controllable, high-entropy sampling engine that extracts diverse synthetic CoT.

\subsection{Zero-Training Stress Testing for Structured Generation}

Deploying language models into operational software pipelines often requires strict adherence to structured output formats, such as JSON schemas, function calling interfaces, and API payloads. While formal programming languages (e.g., Python in HumanEval) exhibit zero lexical redundancy under Taboo masking, semi-structured outputs often contain structural and semantic flexibility, such as key-value ordering, alternative parameter names, or free-text field values. Decoding-level logit masking provides a zero-training, runtime stress-testing framework for these interfaces. By dynamically masking preferred top tokens within allowed schema candidates, developers can probe whether a model genuinely understands underlying schema constraints or relies on fragile, memorized template sequences \cite{ayoobi2026tokenizer}.

\subsection{Improving Off-Path Robustness via Taboo-Guided Alignment}
By integrating Taboo sampling into the rollout phase of verifier-based RL (e.g., GRPO on mathematical domains), optimization can explicitly reward models that discover alternative, valid reasoning paths under forced constraint. This transforms Taboo from a passive diagnostic into an active regularization mechanism, mitigating fragility during policy alignment.

\section{Conclusion}

We introduced Decoding-Level Taboo, a zero-prompt diagnostic that stresses a model's reasoning by masking its preferred tokens at word boundaries during decoding, with injected surprisal as a measured, graded dose. Applied to four open-weight families across four benchmarks, the diagnostic yields a consistent picture: off-path robustness is not a given but an acquired property, post-training alignment confers it (in three of four families), scale compounds it, and the largest aligned models retain most of their multi-step reasoning even under aggressive masking, while base checkpoints degrade far more steeply at every scale. The effect is specific to generative reasoning, invisible to multiple-choice formats, and absent in formal syntax, where a single masked keyword is unrecoverable. Because the intervention is in logit space, Taboo requires no prompt engineering, no retraining, and negligible runtime overhead, making it a practical pre-deployment audit for how a model will behave under real-world constraints, and, potentially, a training signal for making models that stay on their feet when pushed.

\section*{Limitations}

This initial study of decoding-level Taboo has several limitations that are briefly discussed next, outlining how they could be addressed in future works:

\begin{itemize}

    \item \textbf{Scope of Intervention Regimes and Masking Rules:} our current formulation applies a uniform logit mask to the top-$i$ candidates at \emph{every} word start. However, Taboo represents a single instantiation within a much broader design space of decoding interventions. We did not explore sparse or adaptive interventions, such as masking only a subset of word start boundaries, triggering interventions conditionally based on token surprisal thresholds ($\Delta S$), or masking tokens other than the top-$i$ ranks. Systematically mapping this wider landscape of dynamic, targeted, or non-top-$i$ masking rules remains an important direction for future research.
    
    \item \textbf{Language Coverage:} all evaluation tasks in this study were conducted exclusively in English. Because word boundaries, subword tokenization mechanics, and morphological structures vary significantly across typologically diverse languages, our findings on word-boundary logit interventions and recovery kinetics may not directly generalize to non-English or multilingual settings. Therefore, it is of interest to run similar experiments in both high- and low-resource languages.

    \item \textbf{Model Architectures and Tokenizers:} while our empirical evaluations span multiple autoregressive open-weights model families across parameter scaling ladders, our scope is restricted to standard open-weights architectures. Extending dynamic decoding interventions to proprietary closed-source systems, Mixture-of-Experts (MoE) architectures, or multimodal models remains future work. 

    \item \textbf{Decoding regime:} our main experiments use greedy decoding to isolate the direct effect of the logit intervention and to compute exact item-level confidence intervals. To verify that the alignment effect is not an artifact of deterministic decoding, we re-ran the GSM8K sweep under temperature sampling ($\tau\in\{0.7,1.0\}$, top-$p{=}1$, top-$k{=}0$, three seeds each), holding the taboo mask fixed and sampling only the continuation tokens. Measured as marginal retention ($Acc_{\text{taboo}}/Acc_{\text{base}}$), the base-vs-instruct gap, its compounding with scale, and the Llama exception all persist: at $\tau{=}0.7$, retention at $i{=}1$ is $41\%\!\to\!85\%$ for Gemma-3-12B, $53\%\!\to\!91\%$ for Qwen2.5-32B and $38\%\!\to\!86\%$ for Qwen2.5-72B, while Llama-3.1-8B shows no instruct advantage (instruct $\le$ base); at $\tau{=}1.0$ the pattern is unchanged (e.g.\ Qwen2.5-7B $22\%\!\to\!58\%$). Seed-to-seed SD is $\le 0.04$. A full characterization across sampling strategies (top-$p$ nucleus filtering, higher temperatures, and every family$\times$scale cell) remains future work.

    \item \textbf{Task Categorization and Format Sensitivities:} our evaluation suite spans distinct cognitive demands, including multi-step quantitative reasoning, program synthesis, and open-domain factual retrieval. However, evaluating the impact of restricting top token choices on other tasks and benchmarks is needed to better understand its impact.

    \item \textbf{Mixed Precision Constraints:} due to memory constraints at larger parameter scales, larger model regimes were evaluated under low-precision quantization, whereas smaller ladder variants were evaluated in native \texttt{bf16}. Although precision control experiments confirm that quantization noise falls within sampling variance, full high-precision evaluations across all parameter scales are left for future work.

    \item \textbf{Small Family Count and Multiple Comparisons:} the base-vs-instruct comparison spans four families with one family-level exception (Llama-3, absent at 7--8B but reappearing at 70B); we cannot disentangle whether this reflects its specific post-training recipe or an unmodelled confound, and we apply no multiple-comparison correction across the family$\times$size$\times$benchmark$\times$dose grid.
\end{itemize}

\section*{Acknowledgements}
We acknowledge the use of AI assistants for drafting and formatting support during the preparation of this manuscript and to help in the coding of the experiments.

This work was supported by the Agencia Estatal de Investigación (AEI) (doi:10.13039/501100011033) under Grant FUN4DATE (PID2022-136684OB-C22), by TUCAN6-CM (TEC-2024/COM460) funded by CM (ORDEN 5696/2024) and by European Union’s Horizon Europe research and innovation programme under project BRIDGE-AI (Grant 101299050). Access to the Gemini model was provided by the Google Cloud Research Credits program for the Gemini Academic Program. The open-weight models were run on the IPTC computing cluster and on GPUs donated by NVIDIA.

\bibliography{custom}

\appendix

\section{Qualitative Failure Modes under Taboo}
\label{app:failures}

To illustrate how off-path stress degrades reasoning, we inspected the Llama-3.1-8B-Instruct outputs on the $64$ GSM8K items it solves correctly at baseline but fails under Taboo at $i{=}1$. The failures are rarely a single arithmetic slip; more often the model \emph{degenerates}, losing the ability to terminate a coherent chain of thought once its preferred word-initial tokens are blocked. Three recurring modes appear: \textbf{derailment} (no final \texttt{Answer:} line is ever emitted), \textbf{repetition loops} (a short fragment repeats until the token budget is exhausted), and, less often, a confidently \textbf{wrong final answer}. Two representative cases follow; the baseline output is abridged and the taboo output is the model's verbatim continuation.

\paragraph{Example 1 (repetition loop, gold $=25$).}
\emph{Baseline (correct):} ``\dots the dog eats in $2/3\cdot30=20$ minutes; the average is $(30+20)/2=25$.'' $\rightarrow$ \texttt{Answer: 25}.
\emph{Taboo ($i{=}1$):} the chain never converges: ``\dots The average is $(2/3)60$. $60/2=30$. $30/2$ is 15. The cat's time was $(2/3)60$. The average is 30. $60/2=30$. $30/2$ is 15\dots'' [no \texttt{Answer:} line].

\paragraph{Example 2 (degenerate expansion).}
On another baseline-correct item, blocking the preferred continuation sends the model into a non-terminating additive decomposition instead of a solution: ``The number 216 can also be written in the following ways: $216$, $200{+}16$, $200{+}10{+}6$, $200{+}10{+}5{+}1$, $200{+}10{+}5{+}1{+}0$, $200{+}10{+}5{+}1{+}0{+}0$, \dots'' [continues until the generation budget is exhausted].

These degeneration modes, rather than isolated arithmetic errors, are the behavioural signature of Llama-3's missing base-vs-instruct gap on GSM8K (Section~\ref{sec:base_vs_instruct}) and of its flat injected-surprisal response (Figure~\ref{fig:apx_ds}).

\section{Conditional retention with confidence intervals}
\label{app:results_CI}

\begin{figure*}[t]
  \centering
  \includegraphics[width=\textwidth]{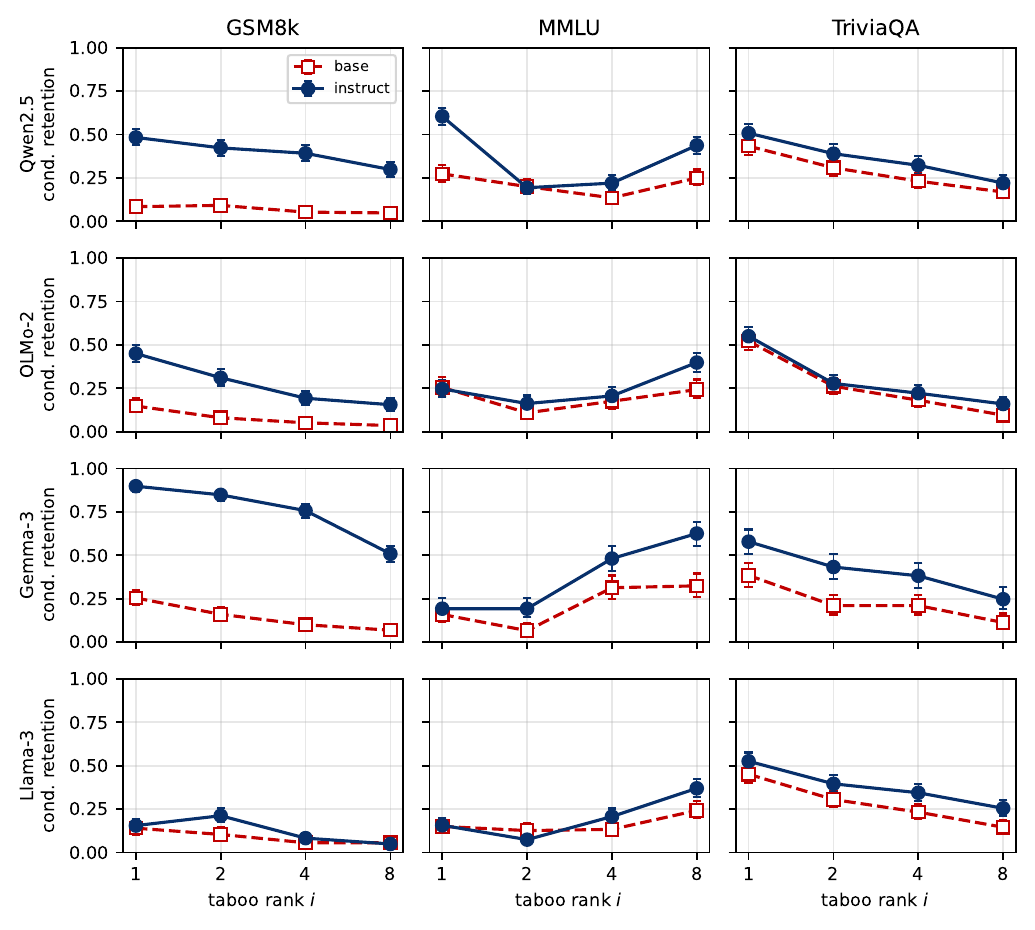}
  \caption{\textbf{Off-path reasoning resilience across model families and benchmarks (representative checkpoints, 7--12B).}
  Conditional retention $\mathcal{R}$ (of items each model answers correctly at baseline,
  the fraction still correct under Taboo) vs.\ taboo rank $i$. Rows: model families
  (\textbf{Row 1} Qwen2.5, \textbf{Row 2} OLMo-2, \textbf{Row 3} Gemma-3, \textbf{Row 4}
  Llama-3). Columns: \textbf{GSM8K} (math reasoning), \textbf{MMLU} (multiple-choice
  knowledge), \textbf{TriviaQA} (factual recall). Dashed = base, solid = instruct, error bars are 95\% Wilson intervals; under
  identical few-shot prompting. On GSM8K the alignment effect is large and family-dependent
  (Qwen/OLMo/Gemma-3) but absent in Llama-3; MMLU is weak and non-monotone (an MCQ-floor
  artifact); TriviaQA gaps are small, consistent with the effect being reasoning-specific.
  HumanEval is omitted as all models collapse to $0\%$ Pass@1 (negative control).}
  \label{fig:main_results_grid}
\end{figure*}

Figure \ref{fig:main_results_grid} shows the conditional retention for the 7--12B model families with 95\% Wilson confidence intervals to visualize the item-sampling uncertainty behind the observed trends.

\section{Absolute accuracy across the full ladder}
\label{app:results_full_lader}

\begin{figure*}[t]
  \centering
  \includegraphics[width=\textwidth]{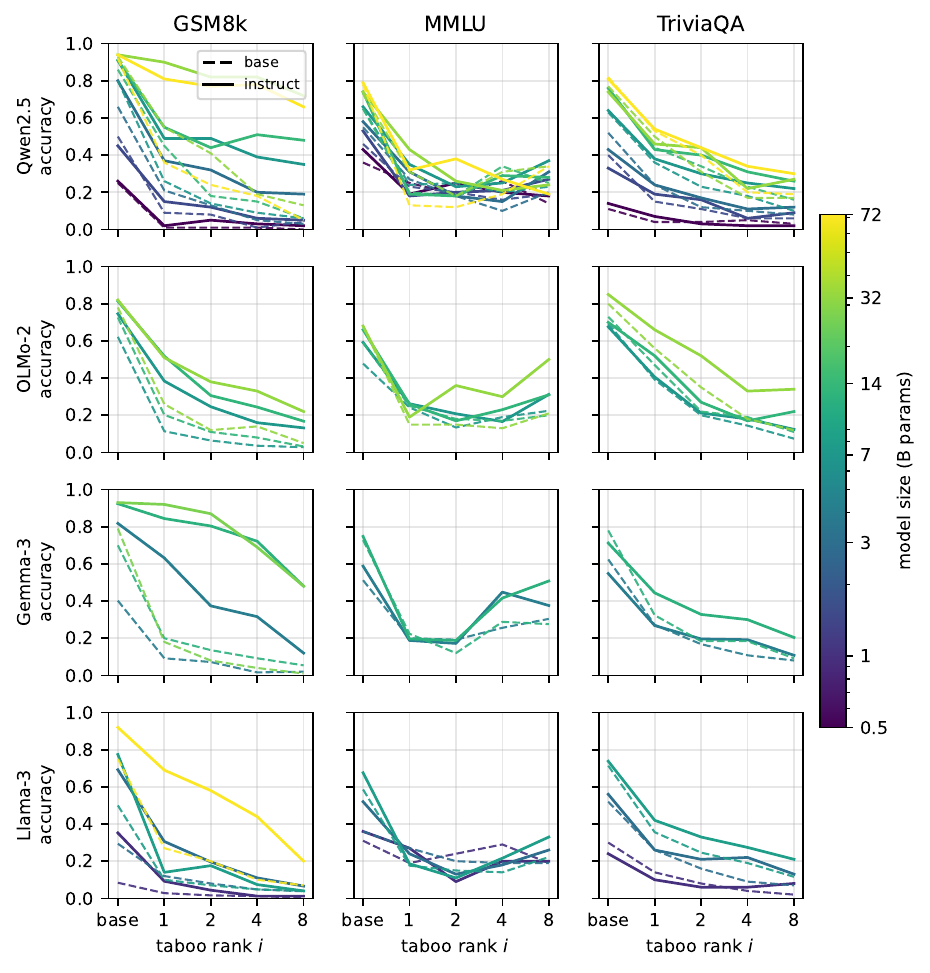}
  \caption{\textbf{Off-path robustness across the full size ladder.} Absolute accuracy vs.\ taboo dose (baseline, then $i=1,2,4,8$) for every model size (viridis: light $=$ small, dark $=$ large), base dashed / instruct solid, across four families (rows) and three benchmarks (columns). We plot absolute accuracy rather than conditional retention here because baseline accuracy varies by an order of magnitude across the ladder, so retention is not comparable across sizes. On GSM8K both scale and alignment lift the curves and compound; MMLU collapses toward the multiple-choice floor at every size; TriviaQA degrades gracefully. At the top of the Llama-3 ladder the instruct curve rises well above base (70B), recovering the alignment gain absent at 8B.}
  \label{fig:allsizes}
\end{figure*}

Figure~\ref{fig:allsizes} presents the absolute accuracy across the entire model size spectrum to provide a comprehensive view of performance under intervention.

\section{Injected surprisal versus model size and type}
\label{app:inj_surprisal}

\begin{figure*}[t]
  \centering
  \includegraphics[width=\textwidth]{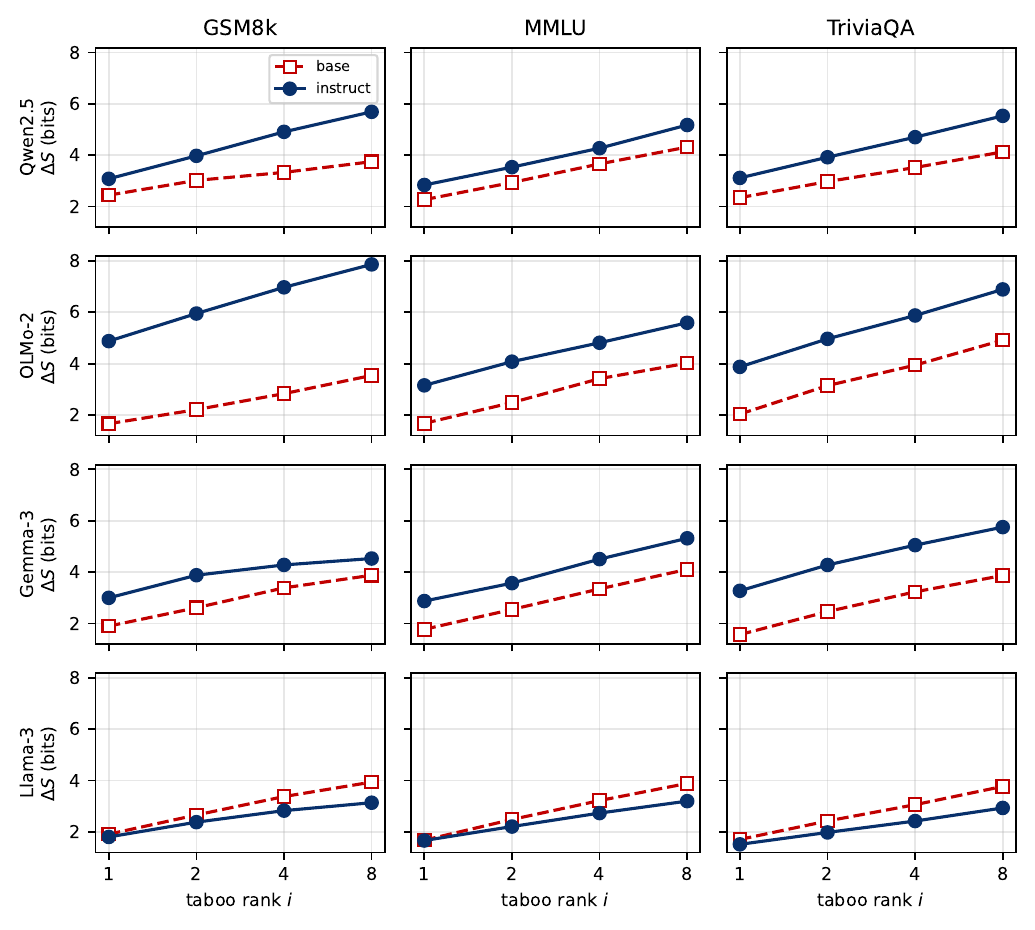}
  \caption{Mean injected surprisal per intervention $\bar{\Delta S}$ (bits) vs.\ taboo rank $i$, base (dashed) vs.\ instruct (solid), for four families at 7--12B (rows) and three benchmarks (columns). Instruct checkpoints absorb a larger dose at every rank in Qwen/OLMo/Gemma but not in Llama-3.}
  \label{fig:apx_ds}
\end{figure*}

Figure~\ref{fig:apx_ds} reports the mean injected surprisal per intervention ($\bar{\Delta S}$) as a function of the taboo rank $i$, for base and instruct checkpoints across the three benchmarks. Two regularities stand out. First, $\bar{\Delta S}$ grows roughly logarithmically with $i$, confirming that the rank parameter behaves as a graded dose. Second, aligned checkpoints absorb \emph{more} injected surprisal than their base counterparts at every rank (their next-token distributions are sharper, so masking the top candidates costs more bits), yet, as shown in the main text, they retain more. The alignment advantage therefore cannot be explained by aligned models receiving a lighter effective dose; they withstand a stronger perturbation and still preserve more of their reasoning. Llama-3 is again the exception: its instruct checkpoint absorbs no more surprisal than its base ($\bar{\Delta S}$ curves overlap or invert), consistent with its missing base--instruct gap on GSM8K.

\section{The Llama Alignment Effect Across Scale}
\label{app:llama_scale}

\begin{figure}[t]
  \centering
  \includegraphics[width=\columnwidth]{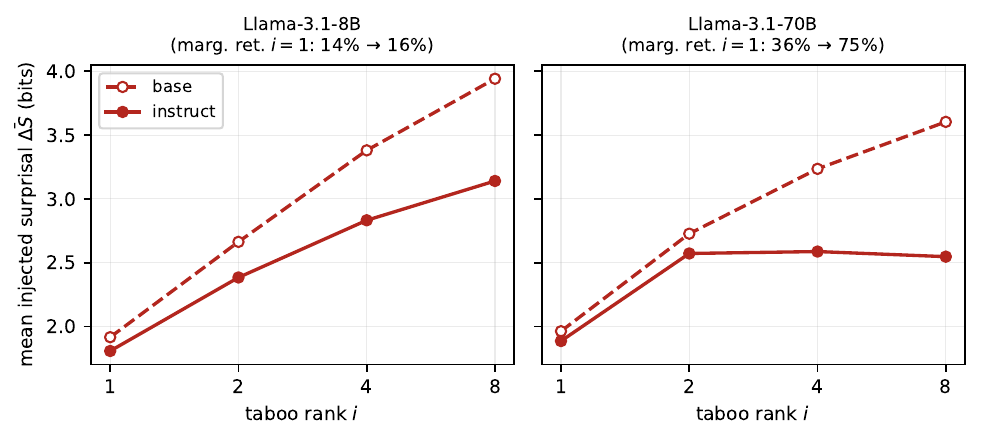}
  \caption{\textbf{The Llama exception across scale.} Mean injected surprisal per intervention $\bar{\Delta S}$ (bits) vs.\ taboo rank $i$ on GSM8K, base (dashed) vs.\ instruct (solid), for Llama-3.1-8B and -70B. At both scales the instruct checkpoint absorbs no more surprisal than its base (slightly less at every rank), so alignment does not sharpen Llama's next-token distribution, unlike Qwen/OLMo/Gemma. Yet marginal retention at $i{=}1$ is flat at 8B ($14\%\!\to\!16\%$) and strongly positive at 70B ($36\%\!\to\!75\%$): the alignment gain at 70B does not pass through distribution sharpening.}
  \label{fig:llama_surprisal}
\end{figure}

In three of the four families (Qwen2.5, OLMo-2, Gemma-3), the alignment benefit under Taboo comes with a clear mechanistic signature: the instruct checkpoint absorbs \emph{more} injected surprisal per intervention than its base (Appendix~\ref{app:inj_surprisal}), consistent with a sharper next-token distribution. Llama-3 is the exception, and Figure~\ref{fig:llama_surprisal} shows this holds at both scales we measured: at 8B and 70B alike, Llama-Instruct absorbs no more injected surprisal than its base, slightly less at every rank, so post-training does not sharpen Llama's next-token distribution the way it does for the other families.

The two scales nonetheless diverge sharply in retention. At 8B the base--instruct gap is flat (marginal GSM8K retention $14\%\!\to\!16\%$ at $i{=}1$), whereas at 70B alignment recovers a large gain ($36\%\!\to\!75\%$). Llama-70B thus acquires taboo-robustness through a route that does \emph{not} pass through distribution sharpening: the surprisal signature is flat at both scales, yet the retention benefit appears only at 70B. This dissociation between the surprisal signature and the retention gain, present in Llama, absent in the other three families, marks the Llama alignment effect as mechanistically distinct. What confers 70B's robustness in the absence of a sharper distribution is an open question our diagnostic makes measurable; a direct measure of distribution sharpness (next-token entropy) is a natural next probe.

Because the 8B and 70B Llama variants utilize an identical tokenizer architecture and vocabulary, the observed behavioral shift across scale cannot be attributed to tokenization mechanics or word-boundary identification artifacts. Instead, it reflects a genuine scaling threshold where the larger latent reasoning engine successfully navigates off-path constraints without relying on distribution sharpening.

\section{Accuracy versus injected surprisal}

\begin{figure*}[t]
  \centering
  \includegraphics[width=\textwidth]{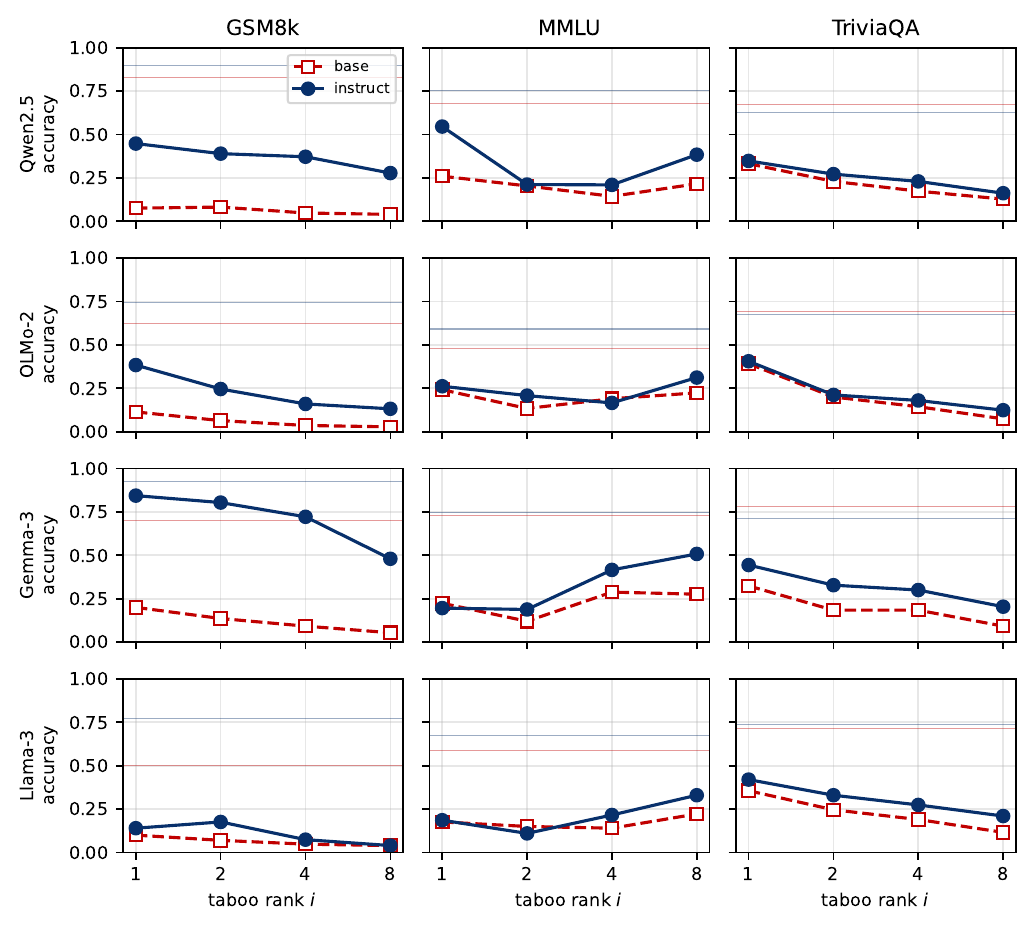}
  \caption{Absolute accuracy vs.\ taboo rank $i$ (thin horizontal lines: unconstrained baselines), base (dashed) vs.\ instruct (solid) for the 7--12B models.}
  \label{fig:apx_acc}
\end{figure*}

\begin{figure*}[t]
  \centering
  \includegraphics[width=\textwidth]{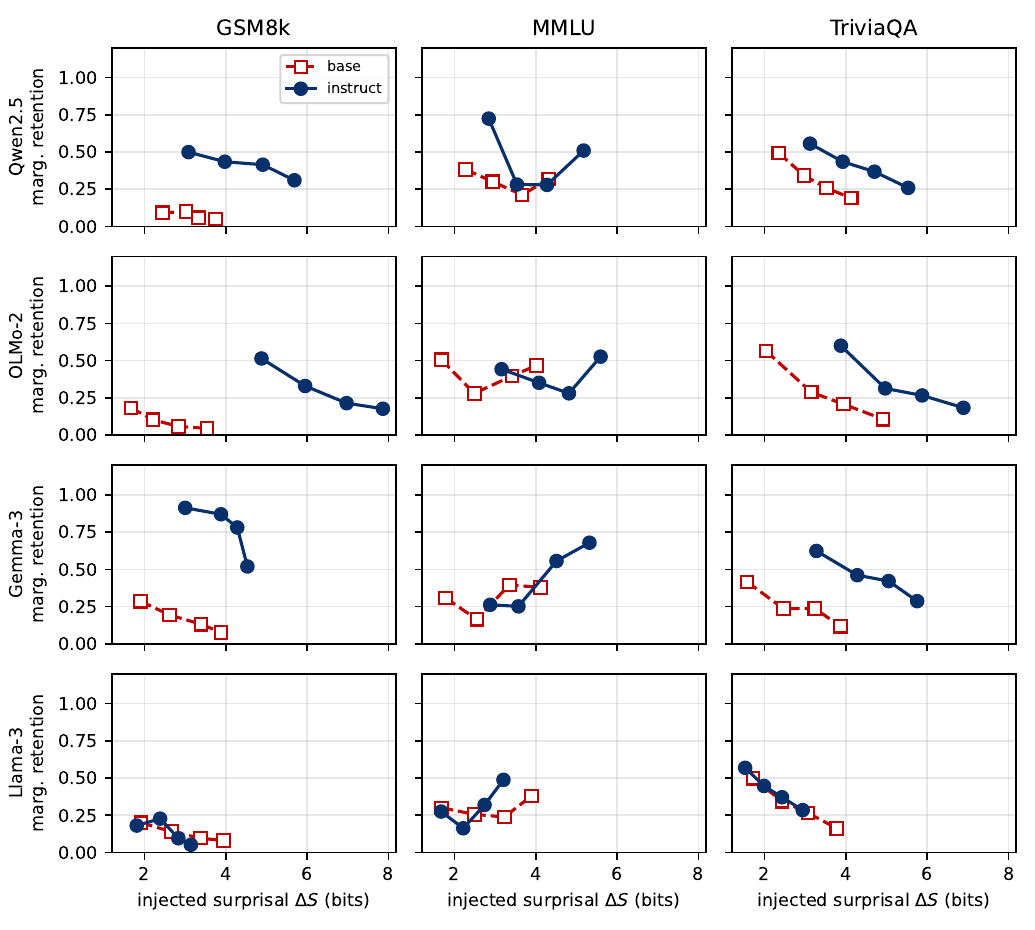}
  \caption{Marginal retention vs.\ measured injected surprisal $\bar{\Delta S}$ for the the 7--12B models. On GSM8K the instruct curves sit above and to the right of base, more retention under a larger absorbed dose, so the alignment effect is not an artifact of a lighter effective perturbation.}
  \label{fig:apx_ret}
\end{figure*}

Figure~\ref{fig:apx_acc} shows absolute accuracy against the taboo rank (with each model's unconstrained baseline as a horizontal reference), and Figure~\ref{fig:apx_ret} re-plots marginal retention ($Acc_{\text{Taboo}}/Acc_{\text{Base}}$) against the measured injected surprisal $\bar{\Delta S}$, i.e.\ against the \emph{effective} dose rather than the nominal rank (here retention is the marginal ratio computed from run summaries; unlike Eq.~5 it also credits items recovered under Taboo, but the two track each other closely). The GSM8K panels make the main-text conclusion explicit in dose space: the instruct curves lie up and to the \emph{right} of the base curves, higher retention at a higher absorbed dose, so the gap is not explained by a lighter effective dose (the curves overlap only partially on the dose axis, so this is indicative rather than a strict dose-match). For Llama-3 the two curves coincide across the measured range.

\section{Answer-Length Expansion under Taboo}
\label{app:token_expansion}

\begin{figure}[t]
  \centering
  \includegraphics[width=\columnwidth]{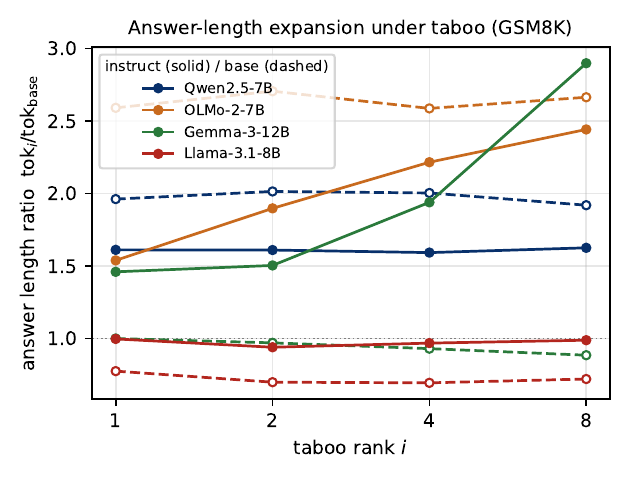}
  \caption{\textbf{Answer-length expansion under Taboo.} Mean generated tokens at dose $i$ divided by the model's own baseline, on GSM8K, for four families at 7--12B (base dashed, instruct solid). Instruct checkpoints spend progressively more tokens as the dose grows (circumlocuting around the blocked continuation): Qwen-7B-Instruct stays near $1.6\times$, while OLMo-2 and Gemma-3 reach $2.4$--$2.9\times$ by $i{=}8$. Ratios at or below $1$ (Gemma/Llama base, Llama instruct) reflect models that already saturate the 512-token budget at baseline or derail rather than expand. For Llama-3.1-8B this is not a truncation artifact: at a 1024-token budget it still saturates at baseline and its taboo retention is unchanged (Section~\ref{sec:cross_family}), i.e.\ it is verbose or non-terminating at baseline, not cut short under Taboo.}
  \label{fig:token_expansion}
\end{figure}

Figure~\ref{fig:token_expansion} illustrates the relative expansion of generated answer lengths as a function of taboo rank, providing empirical evidence of the token overhead associated with circumlocution.

\section{Compute and Reproduction Details}
All experiments ran on a shared cluster with NVIDIA A100 (40\,GB) and H100 (94\,GB) GPUs, one GPU per run. Models up to 14B were run in native \texttt{bf16} or 4-bit; the 32B, 72B, Gemma-3-27B and Llama-3.1-70B models were run in 4-bit (bitsandbytes nf4, bf16 compute) on the H100. We use 4-bit here not for memory, the 94\,GB H100 fits these models in 8-bit, but for speed: \texttt{int8} (LLM.int8) generation ran at only a few tokens per second in our setup, making a full 8-bit sweep impractical, whereas 4-bit nf4 decodes at full speed and matches bf16 within sampling variance (Section~\ref{sec:model_suite}). We therefore use 8-bit only as the spot check on the single outlier cell noted there. Decoding used batch size 8 (4 for the largest models) with generation capped at 512 new tokens.

Per-condition cost is dominated by generation length rather than the intervention itself. Rough wall-clock times on one A100/H100: ${\sim}4$ min per condition for a 7B model at $n{=}100$ (${\sim}20$ min at $n{=}500$), ${\sim}9$ min for 32B, and ${\sim}16$ min for the 70--72B models. A full family/dose sweep thus completes in one overnight session, and the entire study (main grid, size ladder, controls, temperature and precision ablations) is on the order of a few GPU-days on a single accelerator.

\end{document}